%% file: ral2022_gymwyj.tex
\documentclass[letterpaper, 10 pt, conference]{./ieeeconf}

\usepackage{times}
\usepackage[pdftex]{graphicx}
\usepackage{subfigure}
\usepackage{amsmath,amssymb,amsopn,amstext,amsfonts}
\usepackage{cancel}
\usepackage[space]{cite}
\usepackage{pdfsync}
\usepackage{balance}
\usepackage{color}
\usepackage{mathtools}
\usepackage[ruled,linesnumbered]{algorithm2e}
\usepackage{bm}

\usepackage{diagbox}
\usepackage{float}
\usepackage{epstopdf}
\usepackage{pifont}
\usepackage{tabulary}
\usepackage{verbatim}
\usepackage[linkcolor=black,citecolor=black,urlcolor=black,colorlinks=true]{hyperref}
\graphicspath{{./}}
\DeclareGraphicsExtensions{.png,.jpg,.eps,.pdf}
\IEEEoverridecommandlockouts
\title{\LARGE \bf Meeting-Merging-Mission: A Multi-robot Coordinate Framework
for Large-Scale Communication-Limited Exploration}
\author
{Yuman Gao*, Yingjian Wang*, Xingguang Zhong, Tiankai Yang, Mingyang Wang, Zhixiong Xu,\\ Yongchao Wang, Yi Lin\textsuperscript{1}, Chao Xu, and Fei Gao
		\thanks{
	This work was supported in part by the DJI-ZJU FAST Autonomous Drone Research Funding, in part by the National Natrual Science Foundation of China under Grant 62088101.
	Y. Lin\textsuperscript{1} is with Dji Co, Shenzhen, China. 
	E-mail: {\tt\small ylinax@connect.ust.hk.}
	All the other authors are with the State Key Laboratory of Industrial Control Technology, Zhejiang University, Hangzhou, China, and also with the Huzhou Institute of Zhejiang University, HuZhou, China. Corresponding author: Fei Gao.\  
	E-mails:{
			\tt\small 
			\{ymgao, yj\_wang, cxu, fgaoaa\}@zju.edu.cn.
		}
	* Equal contributors.
	}
}

\begin{document}

\maketitle
\thispagestyle{empty}
\pagestyle{empty}

\vspace{-2.2cm}

\input{0_abstract.tex}
\input{1_introduction.tex}

\input{2_related_work.tex}

\input{3_environment_representation.tex}

\input{4_consensus_based_exploration.tex}

\input{5_experiment.tex}
\input{6_conclusion.tex}

\addtolength{\textheight}{1.0cm}

\bibliography{ral2022_gymwyj}

\end{document}

%% file: 0_abstract.tex
\begin{abstract}
\label{sec:abstract}
This letter presents a complete framework Meeting-Merging-Mission for multi-robot exploration under communication restriction. Considering communication is limited in both bandwidth and range in the real world, we propose a lightweight environment presentation method and an efficient cooperative exploration strategy. For lower bandwidth, each robot uses specific polytopes to maintain free space and to generate Super Frontier Information (SFI), which serves as the source for exploration decision-making. To reduce repeated exploration, we develop a mission-based protocol that drives robots to share collected information in stable rendezvous. We also design a complete path planning scheme for both centralized and decentralized cases. To validate that our framework is practical and generic, we present an extensive benchmark and deploy our system into multi-UGV and multi-UAV platforms.

\end{abstract}

%% file: 1_introduction.tex
\section{Introduction}
\label{sec:introduction}

Recently, thanks to the maturity of multi-robot cooperative technology, swarm exploration has received increasing attention in many application areas. Multiple robots can explore wider regions in the time unachievable by a single one, with better fault tolerance and uncertainty compensation. However, in actual exploration missions, communication limitation introduces great challenges to multi-robot exploration tasks and makes the advantages brought by multi robots difficult to leverage. In the real world, especially large-scale environment, it is unrealistic for robots to have global communication capabilities. Besides, transmitting high volumes of sensor data could overwhelm the network capacity. Due to the above realistic factors, the system developed under communication restriction is necessary. 

The communication limitations are considered from the following two aspects: 

(1) \textit{Limited communication bandwidth (LB)}. \textit{LB} makes transmitting the commonly used voxel map or point cloud that are convenient for planning and decision-making exceeds the bearing network capacity.

(2) \textit{Limited communication range (LR)}. Robots are constrained to maintain continuous connectivity or execute tasks lonely, introducing great challenges to exploration.

\begin{figure}[t]
	
	\begin{center}
		\includegraphics[width=0.91\columnwidth]{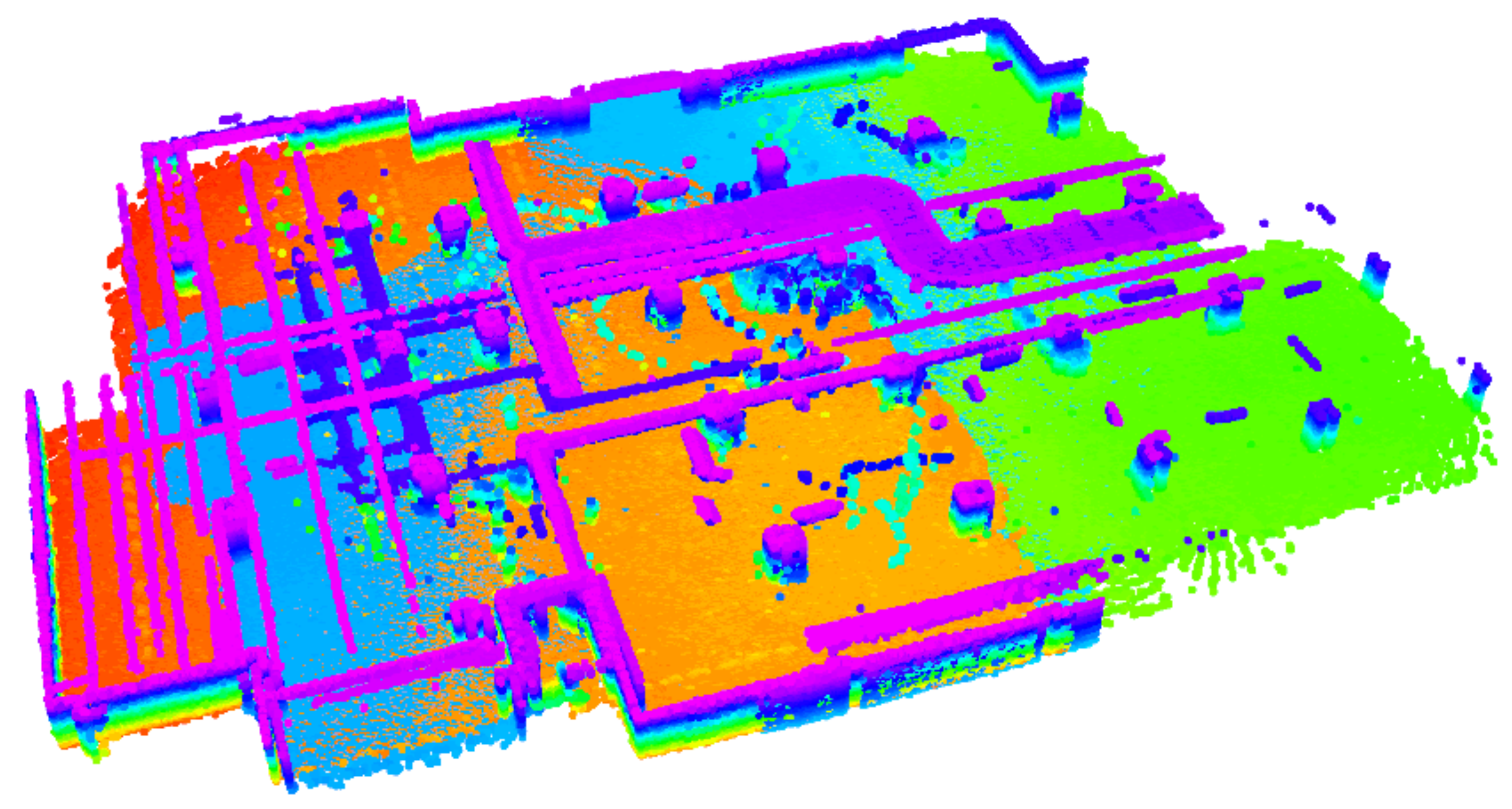}
	\end{center}
	\begin{center}
		\includegraphics[width=0.95\columnwidth]{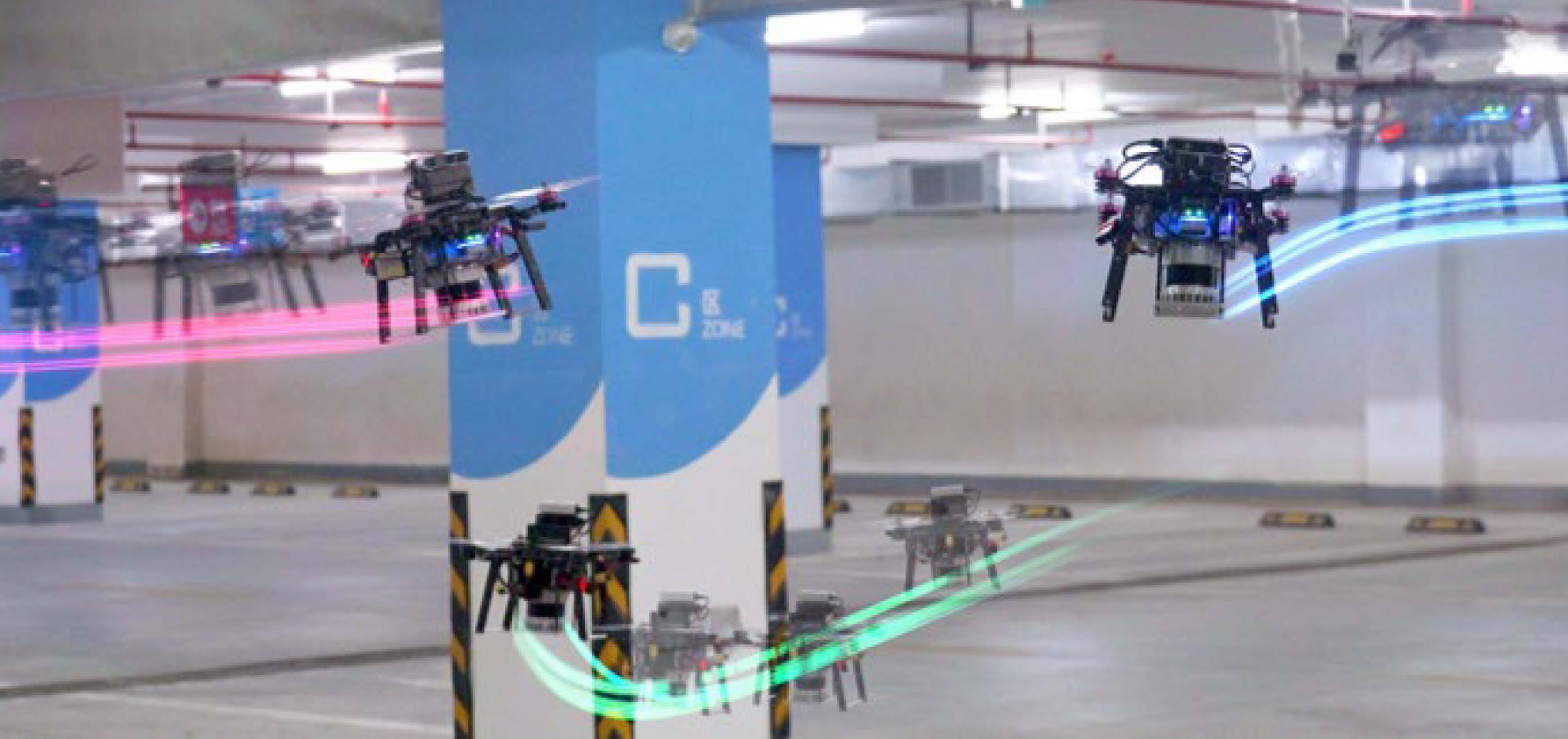}
	\end{center}
	\vspace{-0.2cm}
	\caption{
		\label{fig:head} Composite image of the meeting phase of a multi-robot exploration experiment under communication limit in a large underground parking lot.
	}
	\vspace{-0.6cm}
\end{figure}

\begin{figure*}[t]
	\begin{center}
		\includegraphics[width=1.8\columnwidth]{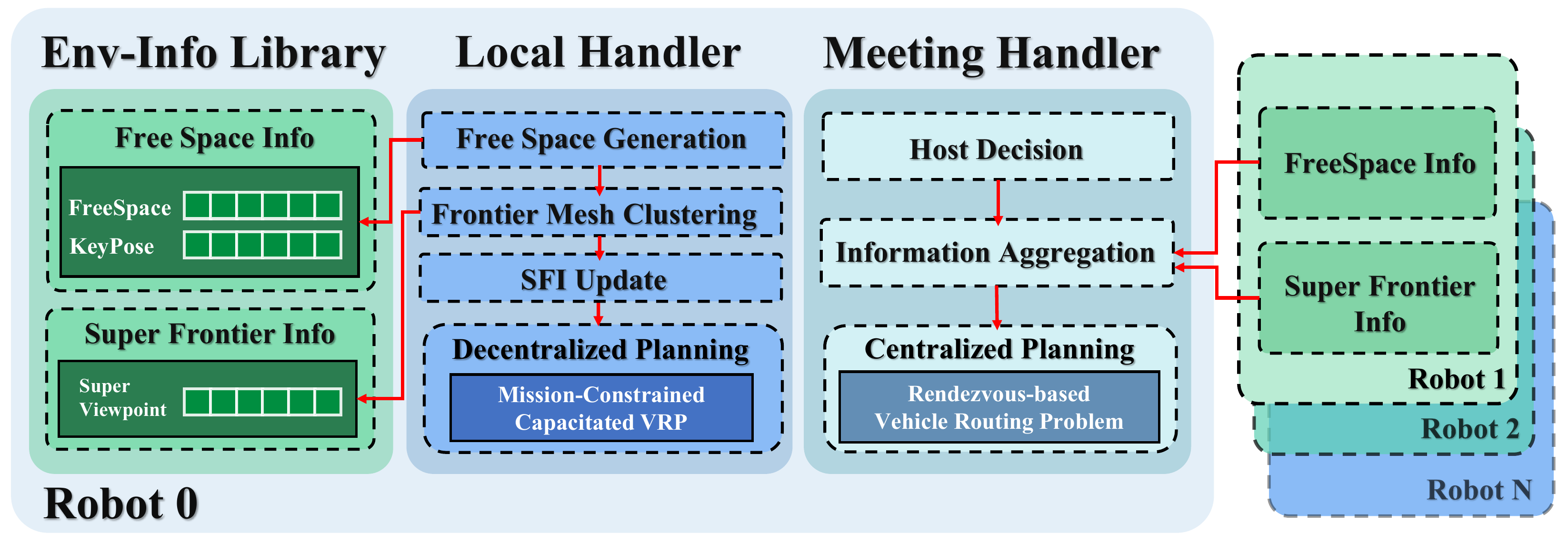}
	\end{center}
	\vspace{-0.5cm}
	\caption{
		\label{fig:framework} An overview of our multi-robot exploration system. Each robot maintain an Environment Library through Local Handler(Sec.\ref{sec:environment representation}). When the connection is built between robots, Meeting Handler will be triggered(Sec.\ref{sec:Mission-based exploration}).
	}
	\vspace{-0.5cm}
\end{figure*}

To resolve the above issues, we propose a complete framework Meeting-Merging-Mission for multi-robot exploration, composed of a lightweight environment presentation method and an efficient cooperative exploration strategy. 

For \textit{LB}, in order to reduce bandwidth for transmission, we use star-convex polytopes to represent known free space. Moreover, utilizing the meshes of the polytopes, we can represent the frontiers which is the boundary of known space. For more efficient exploration decision-making, we generate Super Frontier Infomation (SFI), an integrated information structure representing high-level frontiers and viewpoints. By transmitting star-convex polytopes and SFI, robots obtain the necessary environment information with low bandwidth cost.

For \textit{LR}, we introduce a new mission-based protocol for a team of robots to execute exploration tasks without global communication. The key is assigning missions to robots that guide them to disconnect actively for independent exploration and rendezvous stably for sharing collected information. Besides, we give a complete path planning scheme to balance both exploration mission and requirement of rendezvous in all process of exploration.

Compared with existing state-of-the-art works, our proposed system can explore large-scale environments in less time. We perform comprehensive tests in simulation and real world to validate the efficiency and practicability of our framework. Summarizing our contributions as follows:

(1) A lightweight environment representation using star-convex polytopes and SFI offering essential environment information to drive exploration.

(2) A new mission-based protocol for multi-robot exploration in the absence of global communication. The distributed protocol reduces repeated exploration and increases exploration efficiency.

(3) A complete path planning scheme in all processes of exploration, including centralized planning in joint meeting phase and decentralized planning in lonely exploration phase.

%% file: 2_related_work.tex
\section{Related Work}
\label{sec:related work}

\subsection{Environment Representation}
\label{sec:gmm}
For large-scale scenarios, a lightweight environment representation to is of vital importance to meet the practical communication limit.
Some works \cite{corah2019communication,o2018variable} use the Gaussian mixture model (GMM) as a global spatial representation of the environment.
GMM learns a density function of obstacle point clouds via the expectation-maximization (EM), compresses a huge amount of data as several parameters.  
However, unnecessary computation and inaccuracy have been introduced by GMM, as the free space is not recorded but has to be reconstructed for the component update.
Katz et al. \cite{katz2015visibility} propose to use the HPR (Hidden Point Removal) operator \cite{katz2007direct} to determine the visibility of a point cloud given a viewpoint, without reconstruction or normal estimation. 
Based on HPR, Zhong \cite{zhong2020generating} efficiently generates large, free, and guaranteed convex space among arbitrarily cluttered obstacles.
In this way, the visibility and free space information of a complex environment are extracted by the polytope, which is another compact representation.

The above two groups of representation can both drive robots explore. 
Leveraging the GMM method to model the observed obstacles, information entropy can be calculated for the next viewpoint with large information gain\cite{corah2019communication}. 
Furthermore, the polytope-based method generates free space to distinguish known and unknown regions to drive robot exploration. 
When all the unobserved aims are eliminated by free space, exploration completes.
Yang uses convex polyhedrons to estimate 3D free space in \cite{yanggraph}. However, the convex constraint makes it conservative, especially when robot is in the intersections, reducing the unknown region eliminating efficiency. Williams \cite{williams2020online} uses the method in \cite{katz2007direct} to generate meshes as frontiers. While without maintaining free space, the deletion of frontiers is done by visibility check. The frontier is not visible if there exists another one intersected by the raycast line between the robot and the frontier, which leads the deletion operation to be conservative and not accurate enough, especially when the free space shape is complex.

\begin{figure*}[t]
	\vspace{-0.0cm}
	\begin{center}
		\includegraphics[width=2.0\columnwidth]{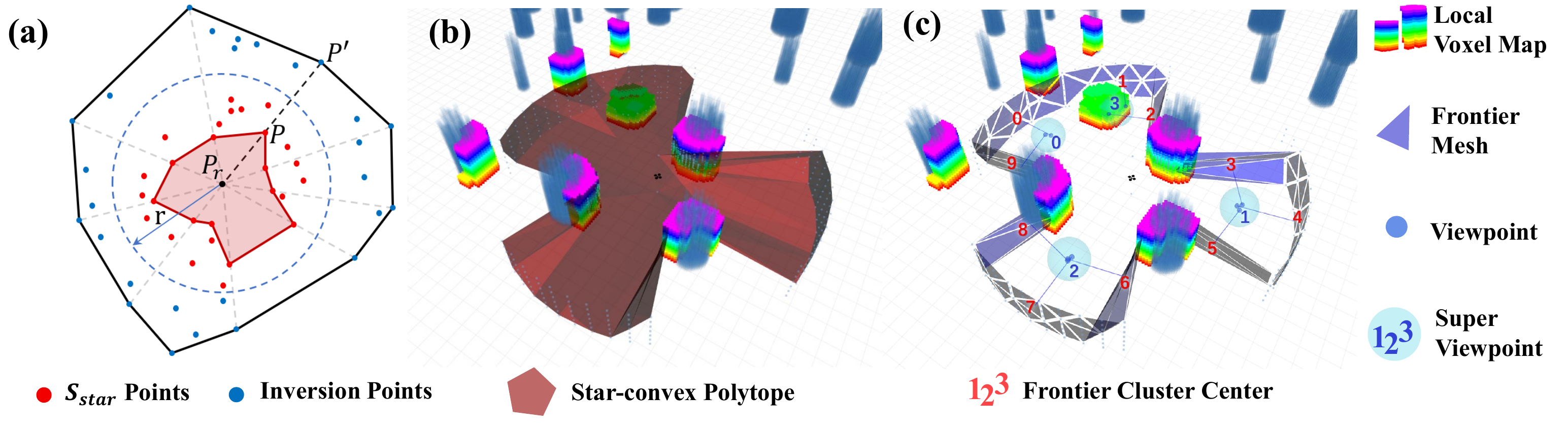}
	\end{center}
	\vspace{-0.4cm}
	\caption{The process of free space and super frontier information generation. (a) Generation process of star-convex polytope. (b) The current free space represented by a star-convex polytope. (c) The SFI of the current frame. Note that cluster $FC_1$ which is marked by red number 1 is generated above a low obstacle showing the spatial representation capability in the z-axis direction.
	}
	\label{fig:freespacegen}
	\vspace{-0.9cm}
	\hspace{10cm}
\end{figure*}

\subsection{Multi-robot Exploration}
Based on the communication mechanism, multi-robot exploration can be summarized into three categories: without any connection requirement, with continuous connection requirement, and with active disconnection and reconnection.

In the first category\cite{burgard2005coordinated, fox2006distributed, zlot2002multi,liu2015leveraging, matignon2012coordinated,andre2016collaboration}, communications are episodic and opportunistic, which could result in repeated exploration and useless energy consumption \cite{amigoni2017multirobot}.
The second category requires robots to keep continuous connection, which is the most restrictive class. 
In \cite{arkin2002line}, robots explore a building subject to the constraint of maintaining line-of-sight communications. 
In \cite{rooker2007multi}, authors present a system in which robots explore the environment while permanently maintaining wireless networking. 
Jensen \cite{jensen2014communication, jensen2018communication} proposes several systems which feature a "mild" form of continuous connection that allows robots to reconnect if it accidentally disconnects
in exploration. 
However, the connection requirements of these approaches might over-constrain the mission objective, resulting in reducted behaviors. 

Besides, the third family of the approaches allows robots to disconnect and reconnect actively.
De Hoog\cite{de2009role, de2010autonomous} innovatively propose a role-based exploration framework, and extend it to cover communication-limited cases. 
Considering the base station, robots are divided into explorers and relays and coordinate through appointed rendezvous positions. The former is assigned to explore unknown environments, and the latter moves back and forth only to deliver information. Later, some work refines the framework. 
Andre \cite{andre2015autonomous} focus on the routing protocols required to share information. 
Cesare \cite{cesare2015multi} presents an interesting feature that UAVs land and act as fixed relays when run out of battery. 
However, even if the role-based framework resolves the limited communication range, the periodic meetings will result in many information-less flights, constraining the exploration process.

%In the field of multi-robot persistent surveillance, robots are also expected to disconnect, monitor and come back. Kantaros \cite{kantaros2016simultaneous,kantaros2016distributed} proposed a system in which robots plan paths to accomplish high-level tasks such as monitoring or transimiting measurement intermittently. In \cite{stump2011multi}, authors are the first to introduce Vehicle Routing Problem(VRP) into field of surveillance and apply this method to the task of surveying a building. In \cite{hollinger2012multirobot} and \cite{banfi2018multirobot}, authors respectively give general characterizations and solutions of Multi-robot Informative Path Planning with Periodic Connectivity (MIPP-PC) and Multi-robot Reconnect Problem(MRP).

Different from existing work, without base-station, our proposed framework considers robots equally. We expect them to disconnect actively for independent exploring but reduce 
information-less flight via our developed mission-based exploration strategy, resulting in efficient 
exploration in large-scale communication-limited environments.

%% file: 3_environment_representation.tex
\section{Environment Representation}
\label{sec:environment representation}
%\subsection{Problem Statemen}
%Given a 3D bounded unknown space $V \subset \mathbb{R}^{3}$, a team of k robots $R=\{r_i,i=1,2,...,k\} $ with communication range limit are deployed to explore the environment. When the distance between robots is greater than the maximum communication distance $L_{com}$ or there are obstacles cutting off the signal, robots will be unable to share information. The region covered by the sensor of robot i $r_i$ is denoted as its known space $V_{known}$. Ideally, it is expected that exploration will lead to $V_{known} = V$. While in real world, some impassable regions and dangerous places such as narrow holes $V_{unexp}$ need to be removed from the expected area of exploration. Therefore, the exploration task completes when $V_{known}=V\setminus V_{unexp}$. 

%Efficient infomation exchange is of vital importance in multi-robot exploration considering communication limit. As defined in [classic frontier], the frontier is the bandary between known and unknown space. And when the frontier is inside current known space(free space), it would be deleted. Then the exploraiton task completes when there is no frontier remains. Therefore, in cooperative exploration, frontiers and free space are the information needed to exchange efficiently.

%Considering communication-efficiency, we use star convex to represent freespace without maintaining the global grid map.

To reduce the bandwidth requirements for transmission, we use the union of a series of star-convex polytopes to represent known free space. We use sampling method to generate star-convex polytope (Sec.\ref{sec:freespace}).
Moreover, we extract meshes from these polytope as frontiers to represent the boundary of known and unknown space. When the free space updates, old frontiers are deleted efficiently (Sec.\ref{sec:sfi_de}).
Then we cluster frontier meshes into frontier clusters (FC) (Sec.\ref{sec:fc}). 
For better observation for robots, we attach a best viewpoint (VP) to each FC and further integrate viewpoints into super viewpoint (SVP) for decision-making (Sec.\ref{sec:vp}). 
All SVPs and included information in them compose super frontier infomation (SFI), as listed in Tab. \ref{tab:notations}.
\begin{table}[b]  
	\centering
	\vspace{-0.4cm}
	\caption{Super Frontier Information}  
	\label{tab:notations}  
	\begin{tabular}{ll}
		\\[-4mm]  
		\hline  
		\hline\\[-2mm]  
		{\bf \small Symbol}& \quad \quad \quad \quad \quad \quad \quad{\bf\small Explanation}\\  
		\hline  
		\vspace{1mm}\\ [-3mm]  
		$F_i$      &  {Frontier mesh with center $c_i$ and normal $n_i$}\\  
		\vspace{1mm}  \\[-3mm]  
		${FC}_j$          &   Frontier Cluster with center $C_j$ and normal ${N_j}$\\ 
		\vspace{1mm}  \\[-3mm]  
		$V\!P_j$          & {Viewpoint of ${FC}_j$}\\  
		\vspace{1mm}  \\[-3mm]  
		${SV\!P}_k$  &   {Super viewpoint}\\  \\[-2mm] 
		\hline   
		\hline  
	\end{tabular}  
	\vspace{-0.0cm}
\end{table}  

\subsection{Star-Convex based Free Space Generation}
\label{sec:freespace}
Star-convex polytope is a specific polytope which can represent known free space by meshes, as shown in Fig.\ref{fig:freespacegen}(b). 

We firstly construct a point set $S_{\text{star}}$ as the source for star-convex polytope generation by sampling in a local voxel map. We uniformly sample points in the cylindrical coordinate system whose origin locates at the position of the robot $P_r$ with radius equals to sensor range $R_{\text{sensor}}$. And the sampling angle range is within robot's field of view. For each sampled point $P_s$, we cast a ray from $P_r$ to $P_s$. If the ray is unobstructed, $P_s$ is added to a point set $S_{\text{free}}$. Otherwise, if the ray hits obstacles, the first point hit obstacles is added to another point set $S_{\text{obs}}$.

Given point set $S_{\text{star}}=S_{\text{free}}\cup S_{\text{obs}}$, we take $P_r$ as origin and use the following sphere mapping function to flip all points in $S_{\text{star}}$ with radius $r$:
\begin{equation}
	P'=F(P)=P-P_r+2(r-\Vert P-P_r\Vert_2) \frac{P-P_r}{\Vert P-P_r\Vert_2}.
\end{equation}
Then a convex hull of the flipped points is calculated and a star-convex polytope is determined by the points on the convex hull after sphere mapping inherently, as shown in Fig.\ref{fig:freespacegen}(a). For more details, we refer readers to our previous work \cite{zhong2020generating}. 
A star-convex polytope is generated when the robot travels a certain distance. The union of a series of star-convex polytopes constitute known free space.

%We then use $S_{star}$, the union of $S_{free}$ and $S_{obs}$, to generate star-convex polytopes based our previous work\cite{zhong2020generating}.

%As shown in Fig.\ref{fig:freespacegen}(a), we uniformly sample points in the cylindrical coordinate system whose origin locates at the position of the robot $P_r$ with proportional decreasing radius $ \{r=r_i |r_i=R_{sensor}/2^i,i\in {0,1,...,n}\}$. 
%Sampling at multiple radius until $r<r_{thr}$ makes free space more accurate near the robot.

%We generate the star convex polytopes using a point set $S_{star}$ sampled in local voxel map, which integrates multi-frame point cloud information.We uniformly sampling points in the cylindrical coordinate system whose origin locates at the position of robot $R_c$ with proportional decreasing radius $ \{r=r_i |r_i=R_{sensor}/2^i,i\in {0,1,...,n}\}$, as shown in Fig.\ref{fig:freespacegen}. Sampling at multiple radius until $r<r_{thr}$ makes free space more accurate near the robot.

%Furthermore, to obtain denser sampling in obstacle-rich regions, we insert points between each pair of adjacent points $P_i,P_j$ recursively, until all sampled points satisfy one of the following conditions:
%\begin{align}\label{equ:mesh}
%	arccos(\frac{(P_j-P_i)\cdot(P_j-P_r)}{||P_j-P_i||_2 ||P_j-P_r||_2}) < \theta_{thr},\\
	%||P_i-P_j||_2 < D_{thr}.
%\end{align}

%for data transmission in later multi-robot coordination.
%After the star convex generation, all vertices constituting the star convex are stored as a free space for data transmission.

\subsection{Frontier Generation and Deletion}
\label{sec:sfi_de}
\begin{figure}[b]
	\vspace{-0.2cm}
	\centering
	\includegraphics[width=1\linewidth]{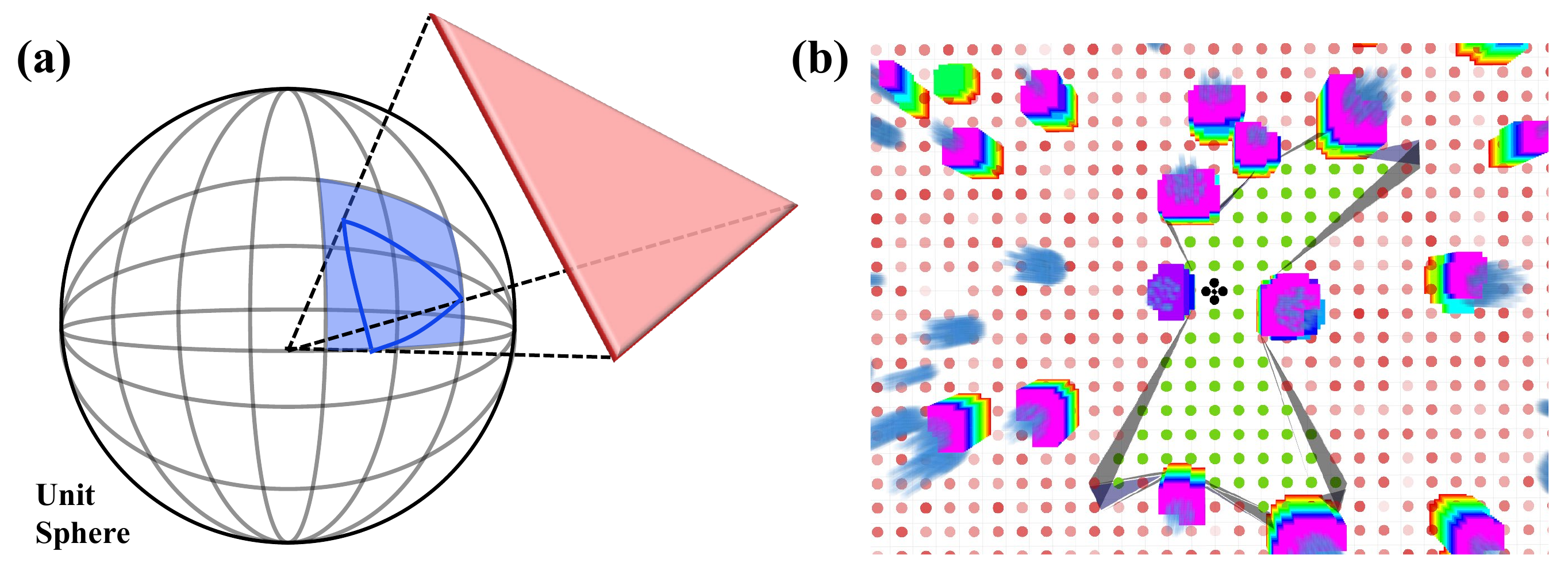}
	\caption{
		(a) We project mesh on a unit rasterized sphere and generate the AABB. (b) The query result. Green and red points are inside and outside a polytope, respectively.
	}
	\label{pic:meshtable}
	\vspace{-0.0cm}
\end{figure}
We represent frontiers of the environment using the meshes of star-convex polytopes. 
To obtain these frontier meshes, we delete meshes whose all vertices belong to $S_{\text{obs}}$ and consider the rest meshes as the set of frontier $F$. 
We denote the purple meshes shown in Fig.\ref{fig:freespacegen}(c) as $F$. For each $F_i$, the center $c_i$ and the normal $n_i$ of it are calculated.  As the normal has two directions, we choose the one satisfying $(P_r-c_i)\cdot n_i > 0$.

When a new star-convex polytope is generated, frontier meshes inside free space should be deleted. 
To this end, we need efficiently query whether a mesh is in a polytope and thus propose \textit{MeshTable} to query if a mesh is inside a star-convex polytope.
As Fig.\ref{pic:meshtable} (a) shows, given a star-convex polytope, we firstly project all its meshes to a unit rasterized sphere. 
For each projected mesh, its axis-aligned bounding box (AABB) on the sphere can be obtained. 
Then each cell in the AABB with its corresponding meshes form a \textit{MeshTable}. In other words, the \textit{MeshTable} records which projected mesh each grid is covered by.

For a mesh $F_i$ to be queried, we project its center $c_i$ to the unit sphere and get the corresponding cell. Then, using the \textit{MeshTable}, meshes corresponded to this cell can be retrieved.
We connect vertices of each mesh with the origin of the polytope to formulate a tetrahedron. 
If $c_i$ is inside one of these tetrahedrons, $F_i$ is inside a star-convex polytope.
A mesh is classified as lying outside free space and deleted if it is not inside any star-convex polytopes.

To speed up the query, we build a KD-tree of all star-convex polytopes' origins.
%to obtain all corresponding to the query mesh $F_i$. 
Then, we search within a radius $R$ centered at $c_i$ using this KD-tree, and obtain corresponding polytopes of $F_i$. 
%Here $R=R_{sensor}+d_{t,thr}$ and $d_{t,thr}$ is a preset maximum distance between two meshes in the same $FC$.
If $F_i$ is judged as inside one of these polytopes using the above-mentioned \textit{MeshTable} query, we delete it.
The result of querying is shown in Fig. \ref{pic:meshtable} (b).

\subsection{Frontier Mesh Clustering}
\label{sec:fc}

\newcommand{\tabincell}[2]{
}

To reduce the number of meshes for efficient decision-making, we cluster the frontier meshes. 
We consider the similarity between meshes from the following three aspects:
\begin{enumerate}
	\item Tangential distance: $d_t(F_i,F_j)=\Vert  (c_i-c_j) \cdot n_j  \Vert_2$,
	\item Normal distance: $d_n(F_i,F_j)=\Vert  (c_i-c_j) \times n_j  \Vert_2$,
	\item Normal difference: $\delta_{n}(F_i,F_j) = \Vert  (n_i-n_j) / 2  \Vert_2$,
\end{enumerate}
where $c_i,c_j$ and $n_i,n_j$ are the center and normal vector of $F_i$ and $F_j$, respectively.
The above similarity criteria are hard to be described by a vector in N-dimensional Euclidean space requiring by most of cluster methods like K-means. So we choose spectral clustering\cite{von2007tutorial}, which only needs the similarity matrix between the data.

For spectral clustering, we need calculate a degree matrix $D$ and a similarity matrix $S$ firstly.
To obtaion $D$, we connect meshes with their k-nearest euclidean-distance neighbors to form a graph, then the degree matrix of the graph is $D$. 
For $S$, based on the above criteria, we have:
\begin{align}
	S=\left(exp(-s(F_i,F_j)^2 / 2\sigma^2)\right)_{i, j=1 \ldots J }, \\
	s(F_i,F_j)=\omega_1 d_t + \omega_2 d_n+\omega_3\delta_{n} ,
\end{align}
where $s(F_i,F_j)$ is the weighted sum of above three distance and $\sigma$ is the preset parameter of Gaussian function. 
Given $D$ and $S$, We can finally get frontier clusters using spectral clustering. 
The clustering example is shown in Fig.\ref{fig:freespacegen}(c), where red numbers represent clusters of frontiers, and the positions of the numbers represent the center of clusters.

\begin{algorithm}[t]
	\label{alg:sfi}
	\caption{Viewpoint Generation} 
	\KwIn{ Frontier cluters $FC_i$ contains frontier meshes $\{F_j,j=1,2,...,J\}$.}  
	\KwOut{Viewpoints $V\!P_i$.}

		$N_{i}=\sum_{F_j \in FC_i}n_j/J$\;
		$C_{i}=\sum_{F_j \in FC_i}c_j/J$\;
		$\mathbf{V}=\mathbf{cylindricalSample}(N_{i}, C_i)$\;
		$s_{best} = 0$\;

		\ForEach{$v_k \in \mathbf{V}$}{
			$d\theta = \mathbf{acos}(\frac{(s_k-C_i)\cdot N_i}{||s_k-C_i||_2})$\;
			$dR = \mathbf{abs}(||s_k-C_i||_2 - R_{opt})$\;
			$score_k = \omega_{\theta}\cdot d\theta+\omega_{R}\cdot dR$\;
			\If{$score_k > s_{best}$}{
				$s_{best} = score_k $\;
				${V\!P}_i = v_k$\;
			}
		}

		%${(N_{PCA})}_{3\times 3},{(\lambda_{PCA})}_{3\times 1} \gets \bm{PCA}({FC_j})$\;
		%$\mathbf{N_j} \gets \bm{calNormalMatrix}(n_{ave}, N_{PCA})$\;
		%$\bm{\omega}_j= [1,0,0]^T, R_m=R_{max}$\;
		%${VP}_j \gets \bm{genVP}({FC}_j,N_j)$\;
		%\While{$\bm{isNotInFree}({VP}_j)$}{${VP}_j \gets \bm{genVP}({FC}_j,N_j)$\;}

		%\Repeat{$\bm{isInFreeSpace}({V\!P}_j)$}
		%{
			%Sampling $\omega_{1} \in [0,\omega_{max}]$\;
			
			%\quad \quad \quad \quad$\omega_{2} \in [0,\omega_{max}\cdot %\lambda_{PCA,2}/\lambda_{PCA,1}]$\;
			
		%	\quad \quad \quad \quad$R_{i} \in [R_{min}, R_{max}]$ \;
		%	$\bm{\omega}=[1,\omega_1,\omega_2]^T$\;
		%	${V\!P}_j = C_j + \frac{\mathbf{N_j} \bm{\omega}}{\Vert \mathbf{N_j} \Vert \Vert \bm{\omega} \Vert } \cdot R_i$ \;
		%}

\end{algorithm}

\subsection{Viewpoint and Super Viewpoint Generation}
\label{sec:vp}

To observe frontier cluster at an appropriate angle and distance, we generate the best viewpoint for each FC by the method presented in Algorithm \ref{alg:sfi}.

As Algorithm \ref{alg:sfi} presents, we firstly calculate the normal $N_i$ and center $C_i$ by averaging all the meshes blong to $FC_i$. Then we score all the points sampled from cylindrical coordinates whose origin locates at $C_i$. The point with smaller angle error to $N_i$ and colser to appropriate distance $R_{opt}$ has higher weighted score. An example of generated viewpoints is shown in Fig.\ref{fig:freespacegen}(c).

%As Fig.\ref{pic:vp}(a) shows, each FC is modeled as an ellipsoid whose pose can be described by three mutually orthogonal unit vectors ${[N_{0},N_{1},N_{2}]}$. We denote  $\bm{N_j}={[N_{0},N_{1},N_{2}]}$.
%To get $N_{0}$ of a frontier cluster $FC_j$, we calculate it as the average normal vector $n_{ave}$. 
%Then, by performing principal component analysis (PCA) for $FC_j$, we get a eigenvector matrix ${(N_{PCA})}_{3\times 3}$ and eigenvalue matrix ${(\lambda_{PCA})}_{3\times 1}=[\lambda_{PCA,0},\lambda_{PCA,1},\lambda_{PCA,2}]$ sorted descendingly. Then we project the last two eigenvectors $N_{PCA,1},N_{PCA,2}$ to the ellipsoid plane as $N_{1},N_{2}$. 

%Using $\bm{N_{j}}$ and the sampling radius and angle, the sample region can be determined as Fig.\ref{pic:vp}(b) shows.
%If sampled viewpoint is not in free space, we gradually change the angle and radius from their initial values and generate again untill ${V\!P}_j$ is feasible. 
%An example of generated viewpoints is shown in Fig.\ref{fig:freespacegen}(c).
%Then, viewpoints can be generated as:
%\begin{align}\label{equ:vp}
%	{V\!P}_j = C_j + \frac{\mathbf{N_j} \bm{\omega}_j}{\Vert \mathbf{N_j} \Vert \Vert \bm{\omega}_m \Vert } \cdot R_j,
%\end{align}
%where $\bm{\omega}_j= [1,\omega_{j,1},\omega_{j,2}]^T, \omega_{j,i} \in [-\omega_{min},\omega_{max}]$, $R_j \in [R_{min}, R_{max}]$, $\omega_{j,i}$ and $R_j$ take the discrete value within its value range. If $VP_j$ is not in free space, we gradually change $\bm{\omega}_j$ and $R_j$ from their initial value and generate again.

To further reduce the scale of decision-making problem, for viewpoints contained in the same sphere with a given thresholding radius, we integrate them as a super viewpoint $SV\!P$, as Fig.\ref{fig:freespacegen}(c) shows.
Finally, we get ${SV\!P}$, each of which consists of $J$ frontier clusters $\{FC_j,j=1,2,...,J \}$ with viewpoint $\{V\!P_j,j=1,2,...,J \}$. 
All the new generated part of SFI will be stored in the Environment Library.

%% file: 4_consensus_based_exploration.tex
\section{Mission-based Exploration}
\label{sec:Mission-based exploration}
%In this section, we describe our proposed mission-based protocol for multi-robot exploration. In a word, 
In this section, we describe an efficient multi-robot exploration strategy with a proposed mission-based protocol.
%which expects each robot to explore environments independently without communication but hold consensus to rendezvous stably for sharing information. 
We divide the process of collaborative exploration into two phases: 
\emph{Joint Meeting} and \emph{Lonely Exploration}, corresponding to the Meeting and Local Handler shown in Fig. \ref{fig:framework}. 
%The remainder of this section details individual modules.

\subsection{Mission-based Protocol for Multi-robot exploration}
\label{Mission Update}
We expect robots to move independently for exploring and meet jointly for sharing information, even in the absence of global communication. 
We define each appointed rendezvous as a mission for robots, including meeting position and time. To achieve our expectation, we develop a centralized planner (see Sec.\ref{Centralized Planning}) in phase \emph{Joint Meeting} for mission decision, and further propose a decentralized planner (see Sec.\ref{Distributed Planning}) in phase \emph{Lonely Exploration} for path planning. 

\begin{figure}[t]
	\vspace{-0.0cm}
	\centering
	\includegraphics[width=1\linewidth]{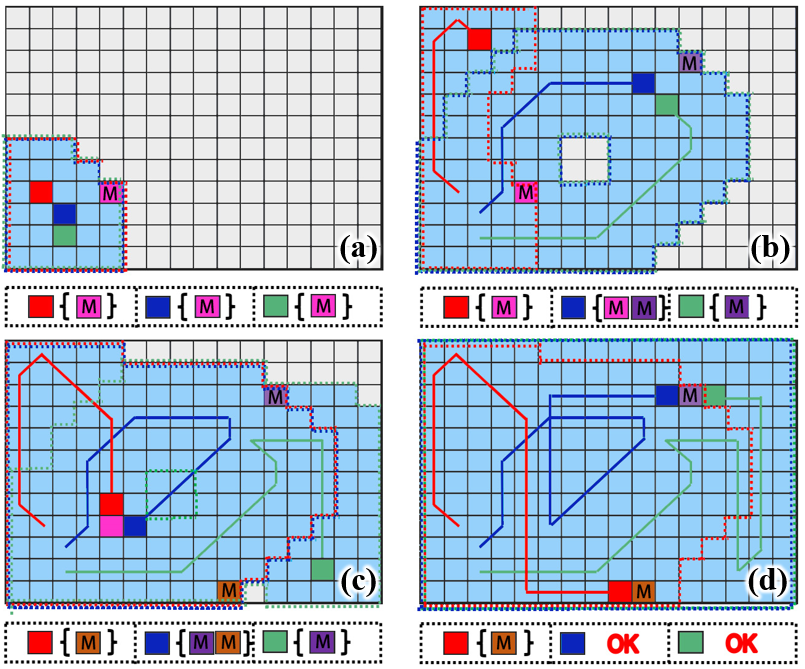}
	\vspace{-0.7cm}
	\caption{The process of a simple demo exploration based on our protocol. Robots are with the colors red, blue, and green, and missions are pink and purple. Grey cells mean unexplored. and light blue cells mean explored.}
	\label{pic:mission update}
	\vspace{-1.0cm}
\end{figure}

At the beginning of each exploration task, robots are assigned a mission in the first meeting, as shown in Fig. \ref{pic:mission update} (a). 
Then, robots spread out to explore independently. 
As the environment explored and new frontiers generated, each robot constantly replans by the decentralized planner, which guarantees the appointed meeting position is arrived on time. 

However, actually, robots may accidentally meet in the \emph{Lonely Exploration} phase. 
For this case, we define an extra rule: if robots with the same mission meet accidentally, they decide a new mission and only one of them keep the old mission. An example is shown in Fig. \ref{pic:mission update} (b), robots (blue and green) with the same mission (pink) meet accidentally and share information. There is no need for all of them to arrive at scheduled position in pink mission. Thus they decide a mission between them (purple) and only let the blue robot meet with the red robot. By this way, the green robot can spend more time on exploration and improve the efficiency.

Based on the above mission-based protocol, robots explore the whole environment by meeting sequentially with limited communication range, as shown in Fig. \ref{pic:mission update} (c) and (d). 

Besides, in each meeting, robots autonomously cooperate to complete information aggregation. All robots send their message, including generated free space information and SFI, to a "host" robot. The host robot merges them and eliminates frontiers that are inside the union of all free space, with similar process to Sec.\ref{sec:sfi_de}. Then the merged map information is send back to all meeting robots.

\begin{comment}
	\subsection{Rule-based Host Decision and Information Aggregation}
	\label{Chairman Election}
	Once meeting, robots need firstly decide a host in charge of information aggregation and centralized decision. 
	Afterward, other participants only send their information to this host.
	In detail, we design three rules for deciding the host robot as follows: 
	\begin{enumerate}
	\item  Robots establishing a new meeting decide the host based on the timestamp of detecting other robots $t_d$. .
	\item When a robot enters a host-decided meeting, it follows other old meeting participants to keep the same host, as shown in Fig. \ref{pic:chairman election} \textcircled{1}. 
	\item If the host leaves a meeting, the rest participants re-decide the host by $t_d$
	as shown in Fig. \ref{pic:chairman election} \textcircled{2}. 
	\end{enumerate}
	
	\begin{figure}[h]
	\vspace{-0.0cm}
	\centering
	\includegraphics[width=1\linewidth]{chairman.pdf}
	\caption{Even if someone enters or leaves a meeting, there is only one recognized chairman in the meeting.}
	\label{pic:chairman election}
	\vspace{-0.3cm}
	\end{figure}
	
	Once the host is fixed, it receives messages from participants, merges them, and sends the merged information back to other robots. 
	The received messages include robots' generated free space information and SFI.
	After merging, some frontiers are inside the union of all free space.
	Therefore, the host eliminates them, with a process similar to Sec. \ref{sec:sfi_de}.
\end{comment}

\subsection{Centralized Optimal Decision Planning}
\label{Centralized Planning}

After information aggregation, by formulating a constrained integer optimization problem, the host decides a mission and assigns it, including the position and time of the next rendezvous. We consider the following rules:
\begin{enumerate}
	\item Robots start from their positions, cross intermediate SVPs and reach a rendezvous position.
	\item Each intermediate SVP is crossed only once. 
\end{enumerate}

To perform optimization, a motion cost between two points is required. It is calculated by:
\begin{equation}
	T_m(p_i, p_j) = \frac{\text{Length}(p_i,p_j)}{v_{\text{max}}},
\end{equation}
where the path length is estimated by A* search on posegraph similar to \cite{lee2021real}, which is not the point of our paper.

We then let $R_c = \{1, ...,n\}$ and $S_c = \{n+1,...,m\}$ denote positions of robots and super viewpoints and define three binary decision variables:
\begin{itemize}
	\item $x^k_{ij}$: set to 1 iff robot $k$ goes from node $i$ to $j$
	\item $y^k_i$: set to 1 iff the node $i$ is crossed by robot $k$
	\item $t_{i}$: set to 1 iff the node $i$ is the rendezvous position
\end{itemize}

The centralized planning problem is formulated as follows:
\begin{align}
\label{equ:5}
\nonumber
&\min_{x^k_{ij}, y^k_i, t_i, \bar{t_i}} J = \sum_{i\in N_c} \sum_{j \in S_c} d_{ij} \sum_{k\in R_c}   x^k_{ij}, \vspace{10ex} \\ 
&~~~s.t. ~ \bar{t_i} = 1 - t_i, ~ \forall i \in S_c, \\
&~~~~~~\sum_{i \in N_c} x^k_{ih} = y^k_h, ~\forall k\in R_c, h\in S_c, \label{eq:1} \\ 
&~~~~~~\sum_{k \in R_c}\sum_{j \in S_c} x^k_{hj} = 1, ~\forall h\in R_c \label{eq:2} \\ 
&~~~~~~\sum_{j \in S_c} x^k_{hj}\bar{t_h} = y^k_h\bar{t_h}, ~\forall k\in R_c, h\in S_c, \label{eq:3} \\
&~~~~~~\sum_{j \in S_c} x^k_{hj} t_h= 0,~ \forall k\in R_c, h\in S_c, \label{eq:4} \\ 
&~~~~~~\sum_{k\in R_c} y^k_h \bar{t_h} = \bar{t_h}, ~ \forall h \in R_c, \label{eq:5} \\ 
&~~~~~~\sum_{k\in R_c} y^k_h t_h = n t_h, ~ \forall h \in R_c, \label{eq:6} \\ 
&~~~~~~\sum_{i \in S_c} t_i = 1, \label{eq:7}
\end{align}
where $N_c = R_c \cup S_c$ and the cost $d_{ij}$ of 
crossing from node $i$ to node $j$ is calculated by $d_{ij} = T_m(p_i, p_j)$. Eq.(\ref{eq:2}) means each robot starts from their current positions. Eq.(\ref{eq:4}) and Eq.(\ref{eq:6}) mean that robots arrive at the viewpoint which is chose as rendezvous position, and Eq.(\ref{eq:3}) and Eq.(\ref{eq:5}) mean that other viewpoints is crossed by a robot once. Eq.(\ref{eq:7}) means that rendezvous position is unique.
\subsection{Hierarchical Sub-optimal Decision Planning}
Actually, routing problem is commonly considered after rendezvous position is fixed. Thus, optimizing rendezvous variables $t_i$ and path planning variables $\{x^k_{ij}, y^k_{i}\}$ jointly is difficult and we plan to take it as future work. In this section, we aim to develop a hierarchical approach by firstly determining the rendezvous position and then solving a simplified problem.
Firstly, we define the distance between a node and robots as 
$$
d(p_v) = \sum_{k\in R_c} T_m(p_v, p_k)
$$
where $p_v$ and $p_k$ is positions of node $v$ and robot $k$. Then we have following options to determine the meeting position:
\begin{enumerate}
	\item \textbf{Furthest-Meeting}: take the node $v = \text{max}_{v \in N_c} d(p_v)$ as the rendezvous position.
	\item \textbf{Nearest-Meeting}: take the node $v = \text{min}_{v \in N_c} d(p_v)$ as the rendezvous position.
	\item \textbf{Shortest-Meeting (Optimal)}: retrieve each node $i$, assume $t_i = 1$ and conduct optimization to get optimal cost $J_i$.  Then select the node $v=\text{min}_{v \in N_c} J_v$ as rendezvous position. 
\end{enumerate}

Each of these methods can simplify formulated problem.
In Sec.\ref{sec:experiment}, we compare the performances of using these methods and choose the \textbf{Furthest-Meeting} for best balancing efficiency and optimality. 

As the rendezvous position is determined, the decision planning problem turns into a vehicle routing problem (VRP) \cite{munari2016generalized}. 
We firstly use a heuristic function for initial path search and then utilize meta-heuristics method for local route search.
In detail, from the positions of robots, we extend paths by iteratively adding the cheapest arc to the routes.
In this way, we obtain an initial solution efficiently. 
Finally, we adopt the extended guided local search (EGLS) algorithm \cite{mills2002extensions} to find an improved solution.

\begin{figure}[h]
	\centering
	\begin{center}
		\includegraphics[width=1.0\columnwidth]{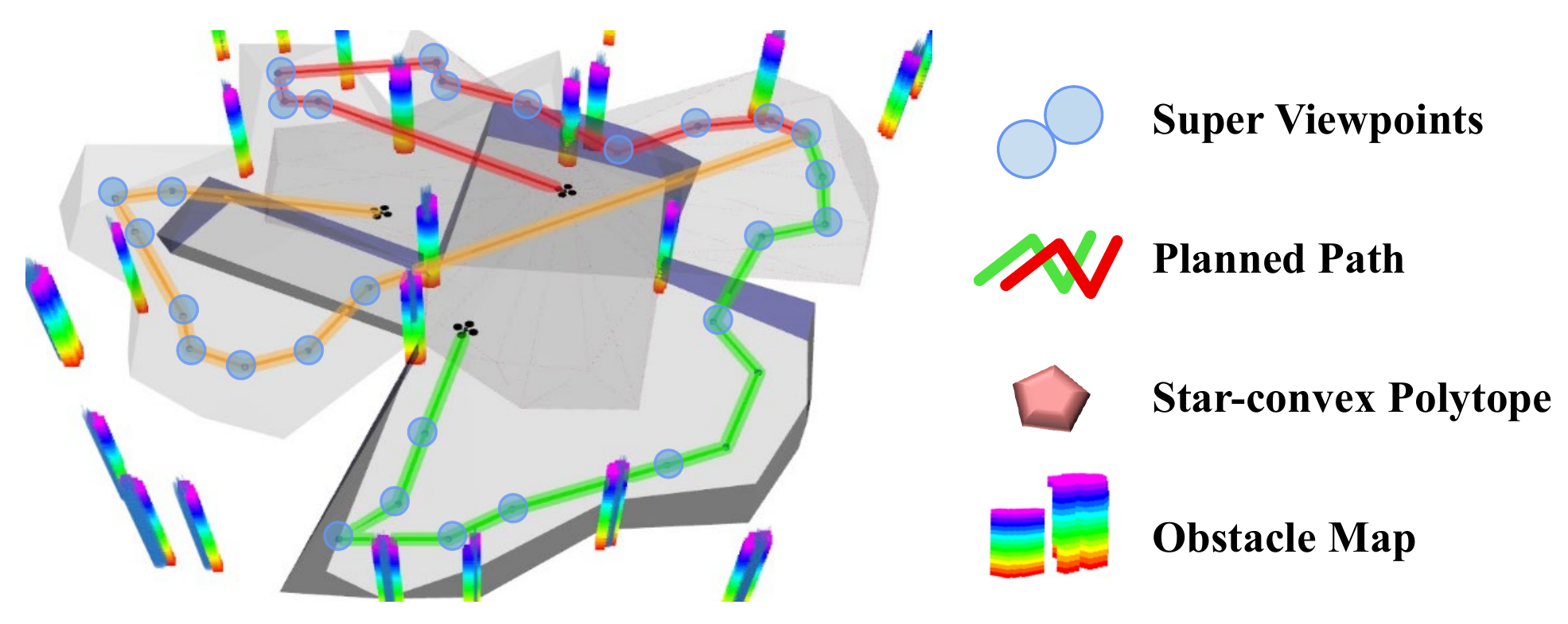}
	\end{center}
	\vspace{-0.4cm}
	\caption{Planned path using the \textbf{Furthest-Meeting} method.}
	\label{fig:3plan}
	\vspace{-0.4cm}
\end{figure}

Until now, we have determined a rendezvous position $P_c$ and some paths $\Psi=\{1,...,n\}$ for robots, where $\Psi^k = \{i|y_i^k=1\}$.
Fig. \ref{fig:3plan} shows the paths by using the \textbf{Furthest-Meeting} method. 
We then choose the maximum cost of paths as the basic time $T_b$ and set $T_e$ as extra time for exploration. 
Besides, to guarantee robots have enough time to rendezvous sequentially, the rendezvous time $T_c$ is:
\begin{align}
    T_c = T_e + \text{max}(T_b + T_{cur}, T^1_l + T_m(P^1_l, P_c), \\
    ..., T^n_l + T_m(P^n_l, P_c)), \nonumber
\end{align}
where $(P^k_l, T^k_l)$ is the last mission of the robot $k$ and $T_{cur}$ is the current time.

\subsection{Decentralized  Path Planning for Single Robot}
\label{Distributed Planning}

After the mission and paths are assigned to robots, they spread out to explore independently. 
Meanwhile, as environment is explored and new SVPs are generated, each robot continuously replans paths to cross some SVPs and arrives at the next rendezvous positions on time.
Furthermore, to alleviate repetitive exploration, we introduce penalties to the area that is assigned to other robots for exploring.

For a single robot $r$, we let $P_r$ be the position of it, $P_m$ be the next appointed meeting position, and $S_d$ be the set of SVPs. 
Then, we define the penalty $p_i$ of node $i$:

$$ p_i = \left\{
\begin{aligned}
\sum_{j \in P_m \cup S_d} d_{ij} &  & (i \in \bigcup_{ k \neq r} \Psi^k) \\
0 &  & (i \notin \bigcup_{ k \neq r} \Psi^k)
\end{aligned}
\right.
$$
where $\Psi^k$ is the path assigned to the robot $k$.

Finally, we formulate the decentralized path planning as
\begin{align}
	\nonumber
	&\min_{x_{ij}, y_i} J = \sum_{i \in N_d} \sum_{j \in P_m \cup S_d} d_{ij} x_{ij} + \sum_{i\in P_m \cup S_d} p_i y_i, \\ 
    &~~s.t.\sum_{i \in N_d} x_{ih} = y_h, ~\forall h\in P_m \cup S_d, \label{eq:2-1}\\
	&~~~~~~\sum_{j \in S_d} x_{hj} = y_h, ~\forall h\in P_r \cup S_d, \label{eq:2-2}\\    
    &~~~~~~\sum_{j \in S_d } x_{hj} = 0,~ h\in P_m, \label{eq:2-3}\\
    &~~~~~~ y_h = 1,~ \forall h \in P_r \cup P_m, \label{eq:2-4}\\
    &~~~~~~\sum_{i \in N_d} \sum_{j \in P_m \cup S_d} d_{ij} x_{ij} \leq T_m - T_{cur}, \label{eq:2-5}
\end{align}
where $N_d = P_r \cup P_m \cup S_d$, $T_m$ is the scheduled time of next meeting, and $T_{cur}$ is the current time. Note that adding penalty $\sum_{i\in P_m \cup S_d} p_i y_i$ is to avoid robot exploring areas that are assigned to other robots. Eq.(\ref{eq:2-1}) and Eq.(\ref{eq:2-2}) provide constrained relationship between $x_{ij}$ and $y_i$. Eq.(\ref{eq:2-3}) denotes that the robot arrives at meeting position and does not leave. Eq.(\ref{eq:2-4}) means that the robot must start from its current position and end with the meeting position. Eq.(\ref{eq:2-5}) means that the robot is guaranteed to arrive meeting in time.

The decentralized path planning problem can be considered as a variant of capacitated vehicle routing problem (CVRP) \cite{munari2016generalized}. 
While a robot moving, it utilizes the latest planned path as an initial solution and refine it using EGLS.

%% file: 5_experiment.tex
\section{Experiment}
\label{sec:experiment}
In this section, we conduct various simulation comparisons and real-world experiments to validate our proposed framework and 
present its advanced performance. 
As for multi-robot autonomous navigation, we use EGO-Swarm \cite{zhou2021ego} to generate smooth and safe tajectories. 

\subsection{Comparisons and Benchmark}
\subsubsection{Bandwidth Comparisons}
\begin{figure}[b]
	\vspace{-0.6cm}
	\centering
	\begin{center}
		\includegraphics[width=1.0\columnwidth]{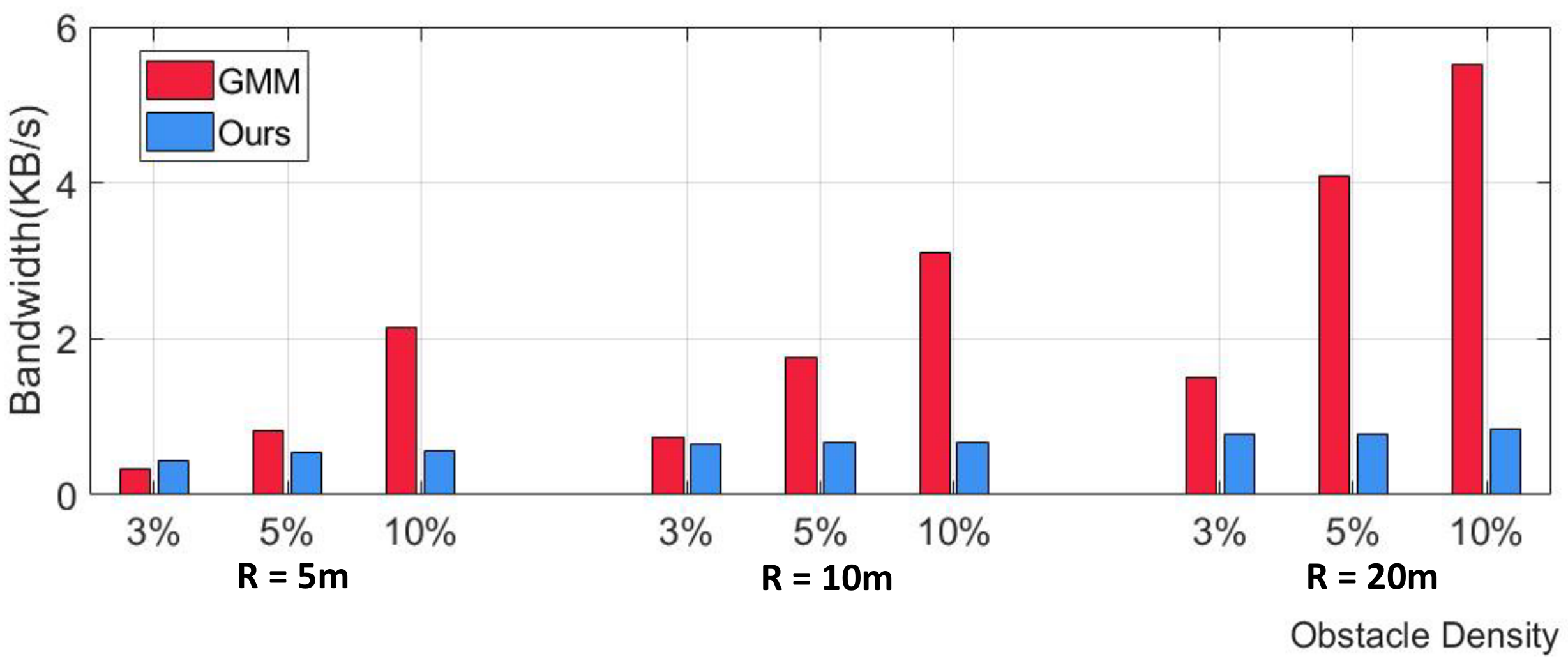}
	\end{center}
	\vspace{-0.6cm}
	\caption{Bandwidth comparision of environment information used to drive exploration using GMM and our method.}
	\label{fig:gmm}
	\vspace{-0.0cm}
\end{figure}

\begin{figure*}[t]
	\vspace{-0.0cm}
	\centering
	\begin{center}
		\includegraphics[width=1.95\columnwidth]{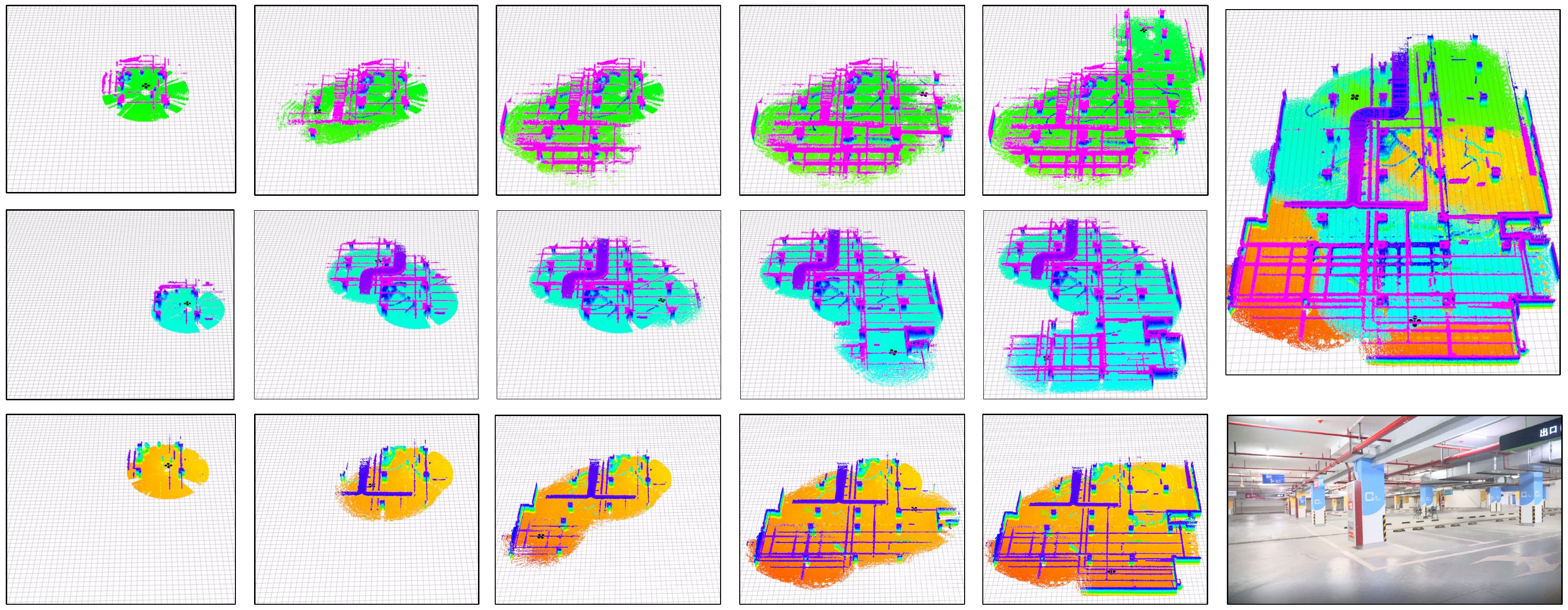}
	\end{center}
	\vspace{-0.4cm}
	\caption{The process of a exploration experiment with three quadrotors. Areas in different colors denote the region explored by different robots. }
	\label{fig:3exp}
	\vspace{-1.4cm}
\end{figure*}
In this part, we compare bandwidth cost with GMM-based methods mentioned in Sec.\ref{sec:gmm}. In GMM method, point cloud $Z$ is modeled as $J$ normal distributions. According to the method and parameters described in \cite{corah2019communication}, we take $J=|\mathcal{Z}| / R_{c}$ with $R_c=160$ to yield good performance. With the same frequency, we compare the bandwidth of data volume of the environment represented by GMM and our method when the data is transmitted over the network. We test with different obstacle densities (percent by volume) and sensor ranges. As the result shown in Fig. \ref{fig:gmm}, the bandwidth cost of our method is lower than the GMM method in all cases, especially for  dense obstacles and large sensor range.

\subsubsection{Meeting Position Decision Comparisons}

\begin{figure}[h]
	\vspace{-0.4cm}
	\centering
	\begin{center}
		\includegraphics[width=1.0\columnwidth]{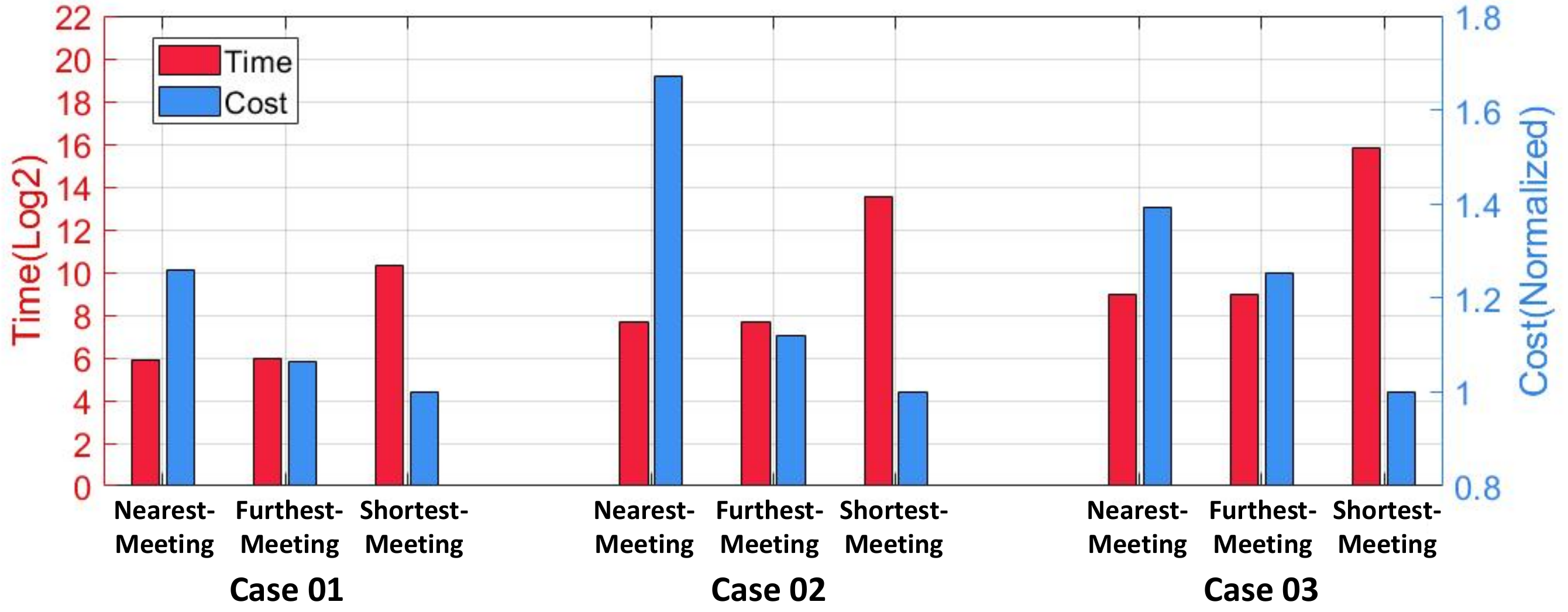}
	\end{center}
	\vspace{-0.5cm}
	\caption{Solving time and cost comparisons between three different meeting place selection strategies.}
	\label{fig:TimeAndCost}
	\vspace{-0.3cm}
\end{figure}

We compare three methods of meeting position selection as mentioned in Sec.\ref{Centralized Planning}. We simulate three cases: three robots with 30-35 SVPs, six robots with 45-60 SVPs, and ten robots with 100-120 SVPs. Each experiment is conducted 20 times in 100m x 100m environments with 100-150 obstacles.
Fig. \ref{fig:TimeAndCost} shows the cost $J$ and solving time $t$ of three methods. As shown, the method with minimum cost consumes the most time, while the \textbf{Furthest-Meeting} method better trade-off between solving time and cost.

\begin{figure}[hb]
	\vspace{-0.4cm}
	\centering
	\begin{center}
		\includegraphics[width=0.85\columnwidth]{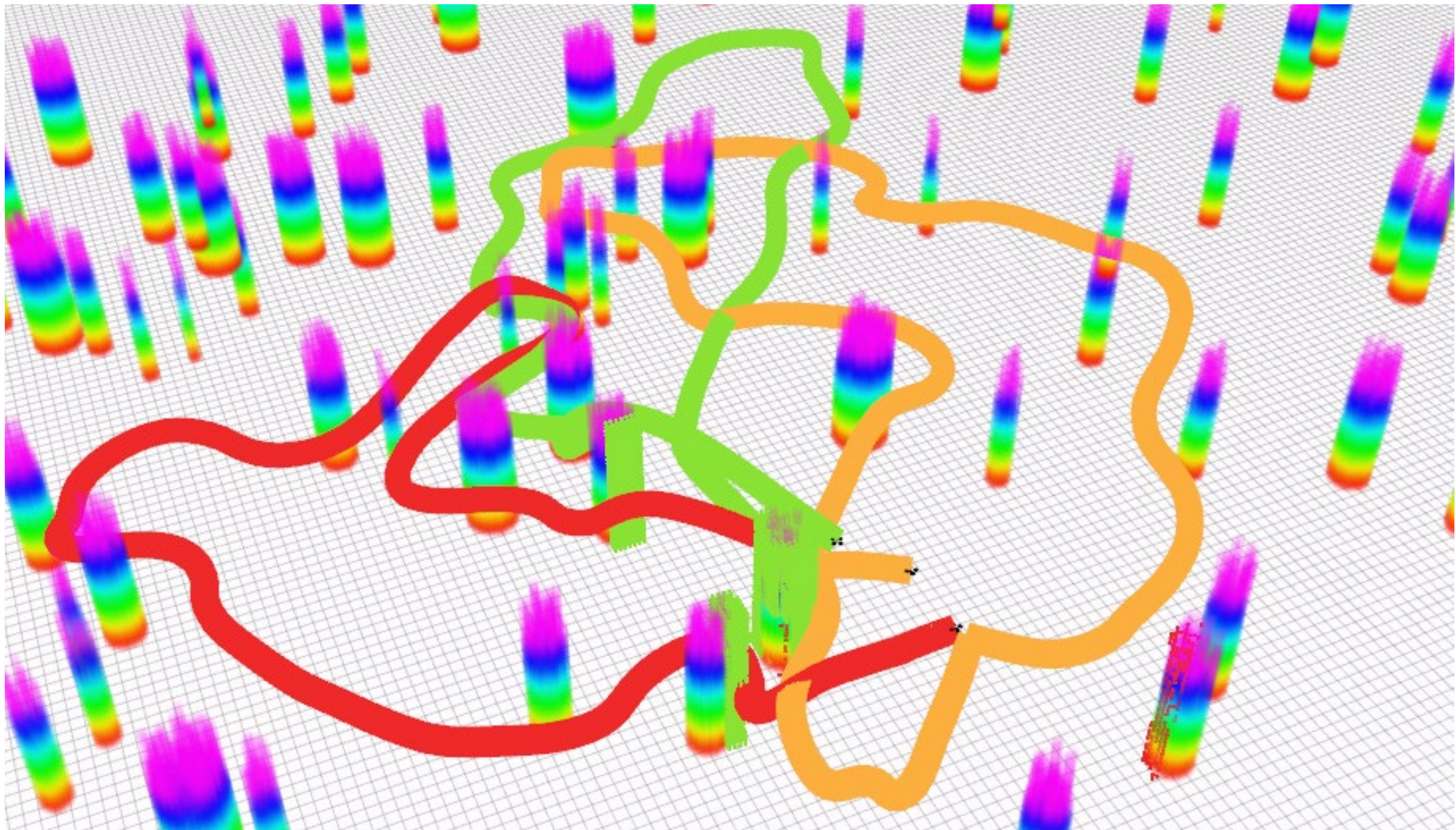}
	\end{center}
	\vspace{-0.4cm}
	\caption{Simulation environment and trajectories of three drones }
	\label{fig:simulation}
	\vspace{-0.2cm}
\end{figure}

\subsubsection{Strategy Benchmark}

We conduct various simulated experiments to compare our method with Burgard's \cite{burgard2005coordinated} and Rooker's \cite{rooker2007multi} methods.
 They are representative works of exploration without communication constraints and with continuous connection requirements, respectively. 
 We simulate several environments with obstacles, shown as Fig. \ref{fig:simulation}. The sensor range and communication range are set to $10m$ and $3m$, respectively.
 We test with different building sizes and robot numbers, with four criteria including exploration time, repeated exploration proportion, independent exploration proportion, and length of trajectories. 
 The results are shown in Tab. \ref{tab:robots_number_benchmark} and Tab. \ref{tab:environment_size_benchmark}. 
 According to the statistics, our proposed method outperforms in exploration time, and efficiently reduces repeated exploration in all cases, especially in large-scale environment.

\begin{figure}[t]
	\vspace{1.0cm}
	\centering
	\begin{center}
		\includegraphics[width=0.7\columnwidth]{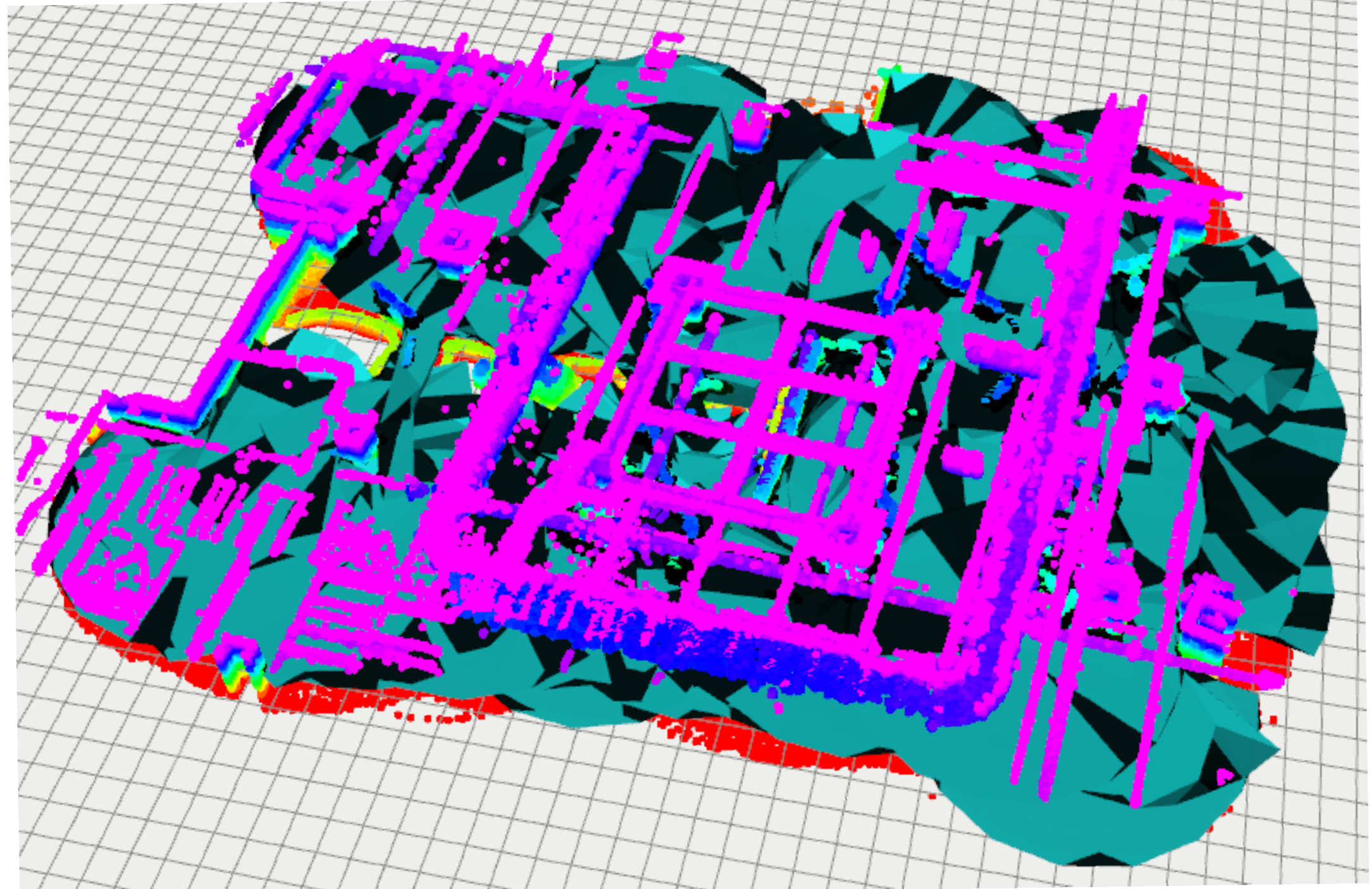}
	\end{center}
	\vspace{-0.4cm}
	\caption{Generated star-convex polytopes in a real-world experiment. }
	\label{fig:realstar}
	\vspace{-2.0cm}
\end{figure}

\begin{table}[ht]
	\vspace{-0.3cm}
	\centering
	\caption{Robots Number Benchmark}
	\setlength{\tabcolsep}{0.5mm}
	\vspace{-0.3cm}
	\renewcommand\arraystretch{1.14}
	{
		\begin{tabular}{|c|c|c|c|c|c|c|}
			\hline
			Scenario    & Method       & time(s) & repeated(\%) & independent(\%) & $l_{traj}$(m)\\  \hline
			
						& Ours      					&  \bf440        & \bf22.2        & \bf61.1   & \bf401    \\ \cline{2-6}
			\#Robots=2	& Burgard's\cite{burgard2005coordinated} & 683       & 63.6      & 80.1   & 663  \\ \cline{2-6}
						&  Rooker's\cite{rooker2007multi} & 1581       & 98.1      & 98.7  & 1439    \\ \hline
						& Ours      					& \bf397       & \bf20.0        & \bf36.5   & \bf351    \\ \cline{2-6}
			\#Robots=3	& Burgard's\cite{burgard2005coordinated} & 492       & 80.9      & 89.7    & 475 \\ \cline{2-6}
						&  Rooker's\cite{rooker2007multi} & 1337       & 96.8       & 95.4   & 1233  \\ \hline
						& Ours      					& \bf403         & \bf20.3          & \bf29.5   & \bf309   \\ \cline{2-6}
			\#Robots=4	& Burgard's\cite{burgard2005coordinated} & 451       & 63.5       & 59.2   & 422   \\ \cline{2-6}
						& Rooker's\cite{rooker2007multi} & 1107       & 95.7       & 94.1  & 907  \\ \hline
	\end{tabular}}
	\label{tab:robots_number_benchmark}
	\vspace{-0.6cm}
\end{table}

\begin{table}[ht]
	\centering
	\vspace{0.3cm}
	\caption{Environment Size Benchmark}
	\setlength{\tabcolsep}{0.5mm}
	\vspace{-0.3cm}
	\renewcommand\arraystretch{1.14}
	{
		\begin{tabular}{|c|c|c|c|c|c|}
			\hline
			Scenario    & Method       & time(s) & repeated(\%) & independent(\%) & $l_{traj}$(m)\\  \hline
			
					& Ours      &  \bf118        & \bf20.1        & \bf60.1     & \bf109   \\ \cline{2-6}
			$2500m^2$	& Burgard's\cite{burgard2005coordinated} & 139       & 63.6      & 81.8     & 131 \\ \cline{2-6}
					&  Rooker's\cite{rooker2007multi} & 128       & 98.1      & 99.2    & 122   \\ \hline
					& Ours      & \bf283        & \bf20.0        & \bf55.3    & \bf271    \\ \cline{2-6}
			$5500m^2$	& Burgard's\cite{burgard2005coordinated} & 398       & 80.9      & 89.2    & 372  \\ \cline{2-6}
					&  Rooker'sk\cite{rooker2007multi} & 801       & 98.2       & 97.7   & 753   \\ \hline
					& Ours      & \bf440         & \bf22.2          & \bf61.1    & \bf401   \\ \cline{2-6}
			$10000m^2$ 	& Burgard's\cite{burgard2005coordinated} & 683       & 70.3       & 79.3    & 663   \\ \cline{2-6}
					&  Rooker's\cite{rooker2007multi} & 1581       & 96.7       & 95.4   & 1439  \\ \hline
	\end{tabular}}
	\label{tab:environment_size_benchmark}
	\vspace{-1.0cm}
\end{table}

\subsection{Real-World Experiment}

\begin{figure}[h]
	\vspace{-0.3cm}
	\centering
	\includegraphics[width=1\linewidth]{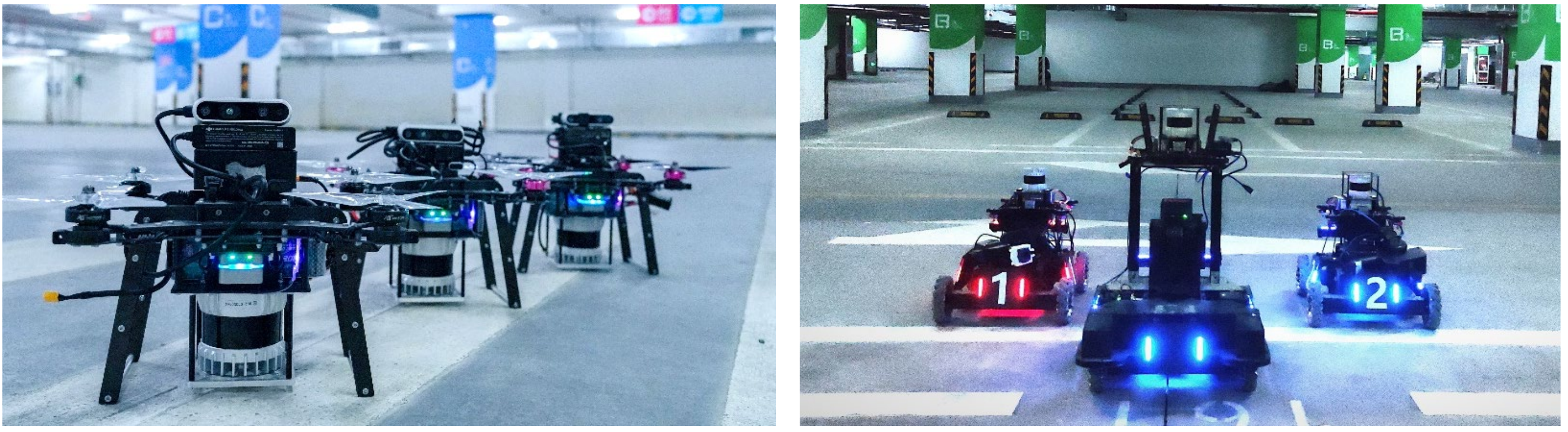}
	\vspace{-0.7cm}
	\caption{Different platforms used for real-world experiments.}
	\label{fig:cars}
	\vspace{-0.5cm}
\end{figure}

Real-world experiments are presented on both UGVs and UAVs platforms, as shown in Fig. \ref{fig:cars}. 
Each of these robots is equipped with a lidar-inertia localization module and a multi-robot planning module. 
They are deployed in a large underground parking lot for exploration.

In the $50m\times 30m$ UGV testing area, we conduct experiments with a 2.5m communication range and a 5m sensor range. 
In the $60m\times 40m$ UAV testing area, we set a 4.5m communication range and an 8m sensor range. 
In all experiments, our proposed framework can drive multi-robot the exploration efficiently under communication limits. 
We refer readers to the video for more information.
As shown in Fig. \ref{fig:realstar}, our generated star-convex polytopes cover the whole explored environment. 
One of the experiment processes is shown in Fig. \ref{fig:3exp}. 
In this experiment, 3 UAVs explore coordinately with a max velocity of 1m/s. 
Even if without global communication, they finish the exploration in 250s. 
For comparison, we also conduct a single UAV exploration.
However, the UAV fails to accomplish the task after it runs out of battery after 8min operation.

%% file: 6_conclusion.tex
\section{Conclusion}
\label{sec:conclusion}
In this paper, we develop a framework for multi-robot exploration under communication limits. To reduce transmission bandwidth, we utilize star-convex polytopes to represent explored free space and incrementally update SFI to drive exploration. To coordinate without global communication, we introduce a mission-based protocol for robots to explore independently and rendezvou to share information. 
%Furthermore, various simulation comparisons and real-world experiments validate that our proposed framework is both robust and efficient.
Future works will be extended to the multi-robot exploration considering localization drift. Robots will plan to actively improve localization quality. Avoiding representing free space via a single occupancy  map, we can conduct loop closures without intractable volumetric map and frontier fusion.